\documentclass[11pt,a4paper]{article}
\usepackage[hyperref]{acl2020}
\usepackage{times}
\usepackage{latexsym}

\usepackage{graphicx}
\usepackage{longtable}
\usepackage{amssymb}
\usepackage{microtype}

\aclfinalcopy 

\title{Aspect Based Sentiment Analysis Using Spectral Temporal Graph Neural Network}

\author{Abir Chakraborty \\
  Microsoft India \\
  \texttt{Abir.Chakraborty@microsoft} \\
   \\}

\date{}

\begin{document}
\vspace{-2mm}
\maketitle
\vspace{-2mm}
\begin{abstract}
The objective of Aspect Based Sentiment Analysis is to capture the sentiment of reviewers associated with different aspects. However, complexity of the review sentences, presence of double negation and specific usage of words found in different domains make it difficult to predict the sentiment accurately and overall a challenging natural language understanding task. While recurrent neural network, attention mechanism and more recently, graph attention based models are prevalent, in this paper we propose graph Fourier transform based network with features created in the spectral domain. While this approach has found considerable success in the forecasting domain, it has not been explored earlier for any natural language processing task. The method relies on creating and learning an underlying graph from the raw data and thereby using the adjacency matrix to shift to the graph Fourier domain. Subsequently, Fourier transform is used to switch to the frequency (spectral) domain where new features are created. These series of transformation proved to be extremely efficient in learning the right representation as we have found that our model achieves the best result on both the SemEval-2014 datasets, i.e., "Laptop" and "Restaurants" domain. Our proposed model also found competitive results on the two other recently proposed datasets from the e-commerce domain.
\end{abstract}

\section{Introduction}
\vspace{-2mm}
With the proliferation of online shopping consumers are relying more and more on ratings, and reviews available from past customers. Aspect-based sentiment analysis (ABSA) tries to understand customers’ granular opinion on different dimensions (aspects) of a product which helps in understanding its current limitations and planning for the next round of improvements. These reviews invariably talk about multiple aspects (sometimes in the same sentence) with the presence of mixed feedbacks, positive and negative. The presence of multiple aspects and opposite sentiments makes the task of ABSA challenging and since its birth in 2014 (SemEval-2014 Task-4, \citeauthor{pontiki-etal-2014-semeval}) the task has seen a variety of different approaches and still enjoys considerable attention from the research community.

The key to an improved performance in ABSA lies in the ability to identify the aspects and the corresponding sentiments with the help of connections between them. A typical example would be “The price is reasonable although the service is poor” where price and service are the aspects with positive and negative sentiments, respectively. It is clear from the sentence that the corresponding modifiers, reasonable and poor, are driving the respective sentiments. However, in another example, “I had the salmon dish and while it was fine, for the price paid, I expected it to have some type of flavor”, it is not clear which words or phrases can be identified as responsible for the negative sentiment. Similar examples are “prices are in line” (neutral sentiment) and “For the price, you can not eat this well in Manhattan” (positive sentiment).
Thus, while one may hope that capturing the syntactical structure and dependency between words will help in improving the sentiment identification it is also important to understand the meaning and application of words in general setting and not to be gleaned from the limited examples present in the modest ABSA datasets (2328 and 3608 training examples for the popular “Laptop” and “Restaurant” domains, respectively). Recent successes with BERT and RoBERTa based encoders indicate that a powerful word representation in the context of the surrounding words would help in establishing the required connections between the aspects and sentiments.

While large model-based encoders provide a powerful initial representation, subsequent transformations possibly involving attention and additional embeddings from parts-of-speech or dependency parser are equally important for the quality of the final prediction. Recent trends also indicate a confluence of deep encoders and graph-based representation of words (nodes) where semantic relations are captured in the graph representations either by converting a dependency graph or by utilizing the nearest neighborhood of words. 

Instead of working on a graph structure stemming from the language itself, a different approach of constructing an underlying graph can be by learning it from the encoded representations. This is motivated by recent advances in the forecasting literature where in absence of any obvious underlying graph, a graph is created from encoded representation by applying self-attention. Armed with this new graph subsequent transformations including Graph Fourier Transform (GFT) and Discrete Fourier Transform (DFT) proved to be extremely efficient in understanding the correlation between different dimensions and increasing the overall performance of the model.   

Motivated by the success of this approach in the forecasting domain, we apply the same for ABSA although we believe there can be many other applications. There are three key components in this technique, (a) create a graph from the encoded representation using self-attention, (b) use the graph Laplacian to transform from the vertex domain to graph-spectral domain and (c) apply DFT to transform from graph-spectral to frequency domain where convolution operator extracts features. To the best of our knowledge these components approach has not been considered in solving any Natural Language applications and would open numerous possibilities for further modifications and improvements. 

The organization of the paper is as follows. In the next section we provide a detailed literature survey on the techniques employed for ABSA. Next, we present the details of the proposed model. Subsequently, the model predictions and comparisons with other baseline methods are discussed. Finally, conclusions are drawn and scope for future works is outlined.

\section{Related Work}
\vspace{-2mm}
Since it’s introduction in 2014 in Sem-Eval (Task-4), (\citeauthor{pontiki-etal-2014-semeval}) ABSA has come a long way from initial SVM classifier with handcrafted features to deep learning classifiers based on RNN, Transformer and memory network. We broadly categorize these models into three groups, (a) RNN with attention, (b) memory network and (c) models that use Transformer architecture and/or pre-trained language model. One of the first application of LSTM was proposed by \cite{wang-etal-2016-attention} where attention mechanism was applied on LSTM output that embedded both the words and aspects individually. \cite{ijcai2018-617} used a hierarchical network of Bi-directional LSTMs with attention at word and phrase level. A new attention model was proposed by He et al. (\citeyear{he-etal-2018-effective}) that improved the performance of the previous LSTM based models. Ma et al. (\citeyear{ijcai2017-568}) proposed an interactive network to learn separate embeddings for the context and target with two sets of LSTMs that attends to specific part of the context based on the target (aspect). Relations between aspects are further investigated by \cite{hazarika-etal-2018-modeling} where hierarchical LSTM structure was used to capture inter-aspect dependency. An attention-over-attention model was proposed by \cite{huang2018aspect} that modelled aspects and sentiments together and explicitly captured their interactions. Similar hierarchical attention model was proposed by \cite{li-etal-2018-hierarchical} which emphasised on the position information of the aspect.

On the models based on memory network, Tang et al. (\citeyear{tang-etal-2016-aspect}) and Chen et al. (\citeyear{chen-etal-2017-recurrent}) designed deep memory networks to with weighted memory mechanism to capture relations between aspects and sentiments separated by long distance. Tay et al. (\citeyear{10.1145/3132847.3132936}) introduced dyadic memory network for ABSA where relevant memory information is adaptively used based on the input query. Cheng et al. (\citeyear{10.1145/3132847.3133037}) proposed hierarchical attention network to have separate aspect attention and sentiment attention that found better matching between previously unseen aspect and sentiment words. Lin et al. (\citeyear{ijcai2019-707}) proposed a new mask memory network with semantic dependency that exploited inter-aspect relations for aspects in the same sentence. 

With the advent of Transformer \citep{vaswani2017attention} and strong baselines reported for different NLP tasks with BERT \citep{devlin-etal-2019-bert} based architectures there are quite a few BERT based models for ABSA. Hoang et al.~(\citeyear{hoang-etal-2019-aspect}) used sentence-pair classification task to reformulate aspect extraction and aspect polarity classification. Zeng et al.~(\citeyear{app9163389}) used BERT embeddings to create a local and global representation of the contexts that were further processed via multi-head self-attention. Xu et al. (\citeyear{xu2019bert}) created a novel task called Review Reading Comprehension from the ABSA datasets and applied BERT to answer the review questions. Li et al.~(\citeyear{li-etal-2019-exploiting}) used BERT as embedding layer together with CRF for end-to-end ABSA. BERT embeddings are also used by Song et al.~(\citeyear{Song_2019}) for ABSA with label smoothing regularization. Sun et al.~(\citeyear{sun2019utilizing}) constructed ABSA as sentence-pair classification task by constructing auxiliary sentences. Phan and Ogunbona (\citeyear{phan-ogunbona-2020-modelling}) combined POS embeddings, dependency embeddings and self-attention with RoBERTa \citep{liu2019roberta} embeddings to further improve on the aspect classification results.


There are not many applications of graph neural networks for ABSA available in the literature. Zhang et al.~(\citeyear{zhang-etal-2019-aspect}) and Sun et al.~(\citeyear{sun-etal-2019-aspect}) used graph convolution network (GCN) where the graph structure was learnt from the dependency tree. Similarly, Huang and Carley~(\citeyear{huang2019syntaxaware}) used graph attention network (GAT) to establish the dependency between words without paying specific attention to the aspects and their opinions. Wang et al.~(\citeyear{wang-etal-2020-relational}) modified the original dependency tree to create an aspect-oriented dependency tree that was used further in a relational GAT (R-GAT) where different relations contributed differently in the computation of nodal representations.

While the above mentioned approaches rely on a graph structure that emerges naturally from the syntactic structure of the examples it is worth exploring if there is a possibility of learning the graph structure itself from the presented data. This idea is borrowed from Forecasting literature where state-of-the-art models are based on GCN originated from the theory of Graph Fourier Transform (GFT). In addition to GCN and temporal modules like LSTM or GRU, it has also been shown that feature processing in the spectral domain can substantially improve the model performance \citep{DBLP:conf/nips/CaoWDZZHTXBTZ20}. While the application of Fourier transform is not common in the natural language processing (NLP) domain, it has been observed recently \citep{leethorp2021fnet} that the self-attention layer in the Transformer can be replaced by a standard Fourier Transform and still achieving 92-97\% of the original accuracy.

\begin{figure*}
\centering
\centerline{\includegraphics[scale=0.5]{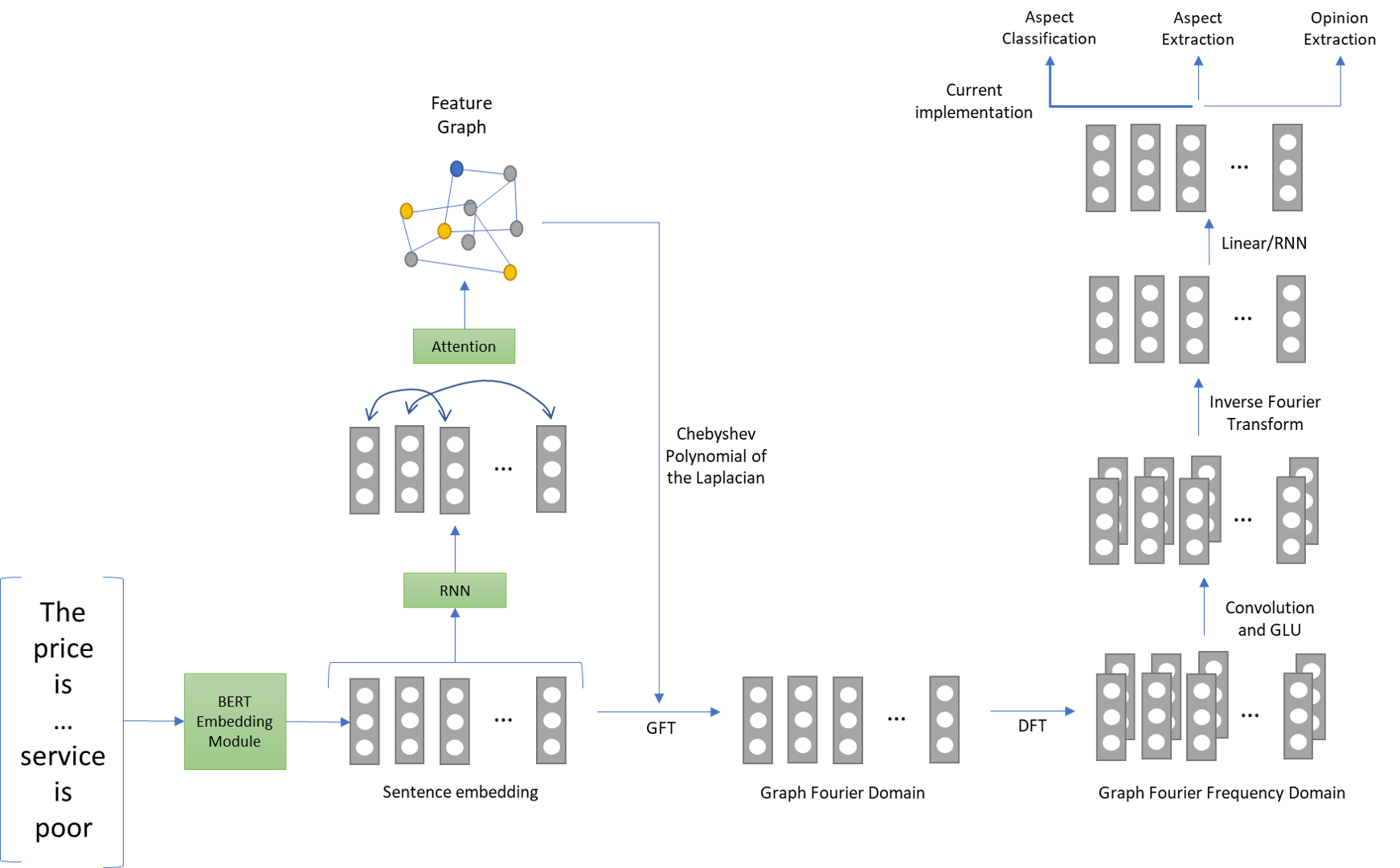}}
\caption{Spectral-Temporal Graph Neural Network model for aspect polarity detection. An example sentence is embedded (using e.g., BERT) first and passed through an RNN and self-attention layer to create a feature graph. The Laplacian of this graph and DFT are used successively to convert the sentence representation into the graph Fourier frequency domain. After extracting features in this domain using 1D convolution and GLU, they are subsequently transformed back to the original domain using inverse Graph Fourier transform and inverse DFT. The final representation is connected to the logits through a fully-connected layer.}
\label{fig:stgnn}
\end{figure*}

\section{Methodology}
The overall architecture of the current method closely follows the architecture of Spectral Temporal Graph Neural Network (STGNN)  \cite{DBLP:conf/nips/CaoWDZZHTXBTZ20} with some minor modifications (see Fig.~\ref{fig:stgnn}). However, for the sake of completeness the components of STGNN are described here. There are five major transformations that any sentence will be subjected to, (1) encoding by an embedding layer, (2) processing by an RNN (we call it the encoder) and create a graph structure, (3) transformation from vertex domain to graph spectral domain using the eigenvectors of this graph, (4) discrete Fourier transform in the graph spectral domain and (5) filtering by convolution layers in the graph spectral frequency domain. Subsequently, inverse Fourier transform and inverse graph Fourier transform are applied sequentially to bring the representation back to the graph vertex domain. These transformations can be broadly combined into three key components, (1) embedding layer, (2) latent correlation layer (LCL) and (3) spectral block layer (SBL). The details of each layer are given below:

\subsection{Embedding Layer}
The embedding layer converts a sentence into a sequence of vectors of some suitable dimension ($h$). Some of the popular choices are (a) Glove vector ($h = 300$), (b) different BERT models (mostly, $h =768$) or RoBERTa models ($h=768$). We have used BERT with 768-dimensional output for all subsequent experiments. While there are different ways of representing a sentence (a) representation of the [CLS] token, (b) average of all the tokens at the last layer or (c) representation of the tokens at the last layer. Here we use the last option where we pass the original sentence along with the aspect term separated by a [SEP] token, i.e., [CLS] + sentence + [SEP] + aspect + [SEP], where + indicates concatenation. Two sentence segments are also created to distinguish between the original sentence and the aspect terms.

\subsection{Latent Correlation Layer}
Given an embedded vector of a sentence LCL learns an underlying graph structure and emits the corresponding graph Laplacian ($L$). This is where STGNN differs from other methods where the graph is computed from the presented data and does not use any external information. First, the encoded representation from the previous layer ($\hat{X} \in \mathbb{R}^{b\times s\times h}$, where $b$ is the batch size and $s$ is the sequence length) is passed through an RNN (of hidden dimension $h$, we have experimented with GRU, LSTM and Bi-LSTM) where the input is posed as a sequence of $h$ elements of dimension $s$. The last hidden state of the RNN ($\breve{X}\in \mathbb{R}^{b\times h\times h}$) is taken as a representation of the entire sentence and passed through an attention layer:
\begin{equation}
W=Softmax\frac{(QK^T}{\sqrt{d}}),\ Q=\breve{X}W^Q,\ K=\breve{X}W^K
\end{equation}
where $Q$ and $K$ can be thought as the query and key learnt from the RNN output through trainable weights $W^Q$ and $W^K$. The output matrix $W \in \mathbb{R}^{h\times h}$ is taken as the weighted adjacency matrix of the graph. 
The adjacency matrix is further processed to create the Laplacian matrix defined as $\mathcal{L} = I_h-D^{-1/2}WD^{1/2}$  where $I_h$ is the $h$-dimensional identity matrix, $D$ is the diagonal degree matrix with $D_{ii}=\sum_{j} W_{ij}$. The eigenvectors of the Laplacian, $U$ (where $\mathcal{L}=U\Lambda U^T$), is used for GFT defined as $GFT(X)=\bar{X} =U^TX$ and inverse-GFT becomes $X=U\bar{X}$. While backpropagation can be applied through eigenvalue decomposition it is often numerically unstable \citep{NEURIPS2019_7dd0240c}. Instead, we apply Chebyshev polynomial approximation \citep{5982158} which only requires Chebyshev polynomials of the Laplacian ($L$) up to a specified order. Thus, if the order of the Chebyshev polynomial considered is $k$ then the GFT is defined as 
\begin{equation}
    \bar{X} = \left[ T_0(\mathcal{L}), T_1(\mathcal{L}), \ldots, T_{k-1}(\mathcal{L}) \right]X,\;\;
\end{equation}
where $T_\ell(\mathcal{L}) \in \mathbb{R}^{N\times N}.$

\begin{table*}
\begin{center}
 \begin{tabular}{|c|c|c|c|c|c|c|c|c|} 
 \hline
 Dataset & \multicolumn{2}{c}{Positive} & \multicolumn{2}{c}{Neutral} & \multicolumn{2}{c|}{Negative} & \multicolumn{2}{c|}{Total} \\ [0.5ex] 
 \hline
  & Train & Test & Train & Test & Train & Test & Train & Test \\
 \hline
 Laptop & 994 & 341 & 464 & 169 & 870 & 128 & 2328 & 638 \\
 \hline
 Restaurants & 2164 & 728 & 637 & 196 & 807 & 196 & 3608 & 1120 \\
 \hline
 Men's Tshirt & 1122 & 270 & 50 & 16 & 699 & 186 & 1871 & 472 \\
 \hline
 Television & 2540 & 618 & 287 & 67 & 919 & 257 & 3746 & 942 \\ [1ex] 
 \hline
 \end{tabular}
\vspace{-2mm}
 \caption{\label{table:datasets}Statistics of the datasets used in this work}
\end{center}
\vspace{-2mm}
\end{table*}


\subsection{Spectral Block Layer}
The transformed representation in the graph Fourier domain is further transformed into frequency domain using DFT. Subsequently 1D convolution followed by a Gated Linear Unit \citep{pmlr-v70-dauphin17a} (originally applied for language modelling) is applied to both the real and imaginary components independently to extract novel features. The output of GLU is transformed back to the time domain using inverse Fourier transform. Subsequently, a linear transformation (akin to inverse GFT) is applied to map back to the vertex domain. Specifically, the DFT output has real and imaginary components, $\hat{X}^r$ and $\hat{X}^i$, that are processed by the same operators (but different parameters) in parallel. The operation can be written as 
\begin{equation}
    M^r(\hat{X}^r) = GLU\left(\theta^r(\hat{X}^r), \theta^r(\hat{X}^r)\right)
\end{equation}
where $\theta^r$ is the convolution kernel of size 3, $\sigma$ denotes sigmoid function, $GLU(x,x) = x \odot \sigma (x)$ and $\odot$ is the element-wise Hadamard product. The same operation is applied to the imaginary components and they are combined together as $M^r(\hat{X}^r) + jM^i(\hat{X}^i)$ ($j^2=-1$) before applying inverse DFT. 

The combined transformation of the LCL and SBL can be thought as another layer that generates an output of dimension same as that of the input which is very similar to the operation of the Transformer layer \cite{vaswani2017attention}, i.e., the output $\tilde{X} \in \mathbb{R}^{b \times s \times h}$. This processed version of the original input $X$ can be transformed further depending upon the nature of the task. For aspect polarity, we explore different options like (1) two fully-connected (FC) layers, (2) GRU followed by a FC layer and (3) LSTM followed by a FC layer. For all these cases, the second FC layer always has two sub-layers with a leaky Relu transfer function in between. The second sub-layer emits raw score of dimension three corresponding to the three sentiment classes (positive, neutral and negative). We use categorical cross-entropy loss with $L_2$ regularization.
The overall time complexity of self-attention and GFT is $O(h^3)$, where $h$ is both the BERT embedding dimension and the hidden dimension of the encoder (GRU/LSTM). The complexity of DFT is $slog(s)$ where $s$ is the sequence length.

\section{Experiments}
In this section, we first describe the datasets used for the evaluation of our proposed method and the other baseline methods employed for comparison. Then, we report the experimental results conducted from different perspectives. Finally, error analysis and discussion are conducted with a few representative examples.

\subsection{Datasets}
We use four public sentiment analysis datasets, two of them are the commonly used Laptop and the Restaurant review datasets from SemEVal-14 task \citep{pontiki-etal-2014-semeval} and other two are recently released and based on e-commerce reviews, namely, Men’s T-shirt and Television (\cite{10.1007/978-3-030-72240-1_7}). Statistics of these datasets are given in Table~\ref{table:datasets}. Looking at the datasets it is apparent that in general we do not have enough training data for most of the deep learning based models and one has to be careful to avoid over-fitting. We also experiment on the "hard-data" as defined by \cite{xue-li-2018-aspect} where examples with multiple aspects and different polarities are identified. 
All experiments were conducted on Tesla K-80 with 12 GB GPU.

\begin{table*}
\begin{center}
\vspace{-2mm}
\begin{tabular}{|c|c|c|c|c|c|c|} 
 \hline
 Model & \multicolumn{2}{c|}{Reported (no held out)} & \multicolumn{2}{c|}{Reproduced (no held out)} & \multicolumn{2}{c|}{Reproduced using 15\% held out} \\ [0.5ex] 
 \hline
  & Accuracy & F1 & Accuracy & F1 & Accuracy & F1 \\
 \hline
 ATAE-LSTM & 68.70 & - & 60.28 	& 44.33	& 58.62 (33.47)	& 43.27 (29.01) \\
 \hline
 RAM & 74.49 & 71.35 & 72.82 & 68.34 & 70.97 (56.04)	& 65.31 (55.81)
 \\
 \hline
 IAN & 72.10 &	-	& 69.94 & 62.84	& 69.40 (48.91)	& 61.98 (48.75) \\ \hline
 BERT-SPC & 78.99 & 75.03 & 78.72 & 74.52 & 77.24 (59.21) & 72.80 (59.44) \\ \hline
 BERT-AEN & 79.93 & 76.31 & 78.65 & 74.26 & 75.71 (46.53) & 70.02 (45.22) \\ \hline
 LCF-BERT & 77.31 & 75.58 & $\mathbf{79.75}$ & $\mathbf{76.10}$ & 77.27 (62.57) & 72.86 (62.71) \\ \hline
 R-GAT+BERT & 78.21	& 74.07 & 79.15 & 75.14 & 75.64 & 69.52 \\ \hline 				
 STGNN-GRU	& - & - & 79.09* & 75.28*  & $\mathbf{78.72 (64.36)}$ & $\mathbf{74.84 (64.34)}$ \\ \hline
 \end{tabular}
 \caption{\label{table:laptop}Comparison of predictions on the Laptop dataset. The best results are highlighted in bold and * next to a number indicates the second best result.}
\vspace{-2mm}
\end{center}
\end{table*}

\subsection{Implementation Details}
\vspace{-2mm}
We extend the codebase of \cite{10.1007/978-3-030-72240-1_7} by adding our proposed model. We have used 768-dimensional embeddings of BERT \citep{devlin-etal-2019-bert} implemented in the PyTorch environment. There are several hyperparameters that we should tune for, namely, learning rate, dropout rate, regularization parameter $L_2$ weights and STGNN specific parameters like the number of layers, encoder and decoder types (fully-connected, GRU, LSTM, Bi-LSTM etc.) etc. However, what we have found is that the optimal set of parameters can be different for different datasets and it would take substantial amount of computational effort to obtain all four of them. 

In this work we have not done an extensive search of the hyper-parameter space. Instead, we started with the baseline parameters used earlier \citep{10.1007/978-3-030-72240-1_7} and modified only the $L_2$ weight that we found to be significantly affecting the test results. Thus, all subsequent results are based on $L_2=2\times 10^{-7}$ whereas the other parameters are as follows: (a) learning rate $=1\times 10^{-5}$, (b) dropout = $0$ and (c) batch size $=32$. We have used Adam optimizer with the default parameters ($\beta_1 = 0.9$ and $\beta_2 = 0.999$) and weight decay. As reported by \cite{10.1007/978-3-030-72240-1_7} most of the earlier studies did not set aside a separate test set and the same dataset was used for validation. However, in this work we follow the same process of keeping 10-15\% of the train data as the validation set. The first pass runs over all the epochs and the optimal epoch number is noted that corresponds to the maximum validation accuracy. Next, the entire training set is considered for training but only up to the optimal epoch and finally the model performance on the test data is reported. 

\subsection{Baseline Methods}
\vspace{-2mm}
We compare with the methods studied by \citep{10.1007/978-3-030-72240-1_7} along with the R-GAT model of \cite{wang-etal-2020-relational}. The methods compared by \citep{10.1007/978-3-030-72240-1_7} can be broadly categorized into two classes, (a) memory network based and (b) BERT based. While memory network based models have fewer parameters and better suited for the small datasets the BERT based methods are dominating the ABSA landscape and their success can be attributed to the huge pre-training corpora that helps in better understanding of words and their associations. A brief description of the methods considered here are given below: 
\vspace{-2mm}
\begin{enumerate}
    \item ATAE-LSTM \citep{wang-etal-2016-attention} where separate embeddings are used for the aspects and concatenated with word embeddings followed by an attention layer. 
    \item Recurrent Attention on Memory (RAM, \cite{chen-etal-2017-recurrent}) where memory network is used to capture relations between aspects and sentiments separated by long distance.
    \item Interactive Attention Network (IAN,  \cite{ijcai2017-568}) where two sets of LSTMs are used to learn the embeddings of the context words and target (aspect). The attention based representations are then concatenated to predict the aspect polarity. 
    \item BERT-SPC, which is a baseline BERT model that treats sentiment classification as a sentence pair classification task where the pooled output of a modified sentence $[CLS] + $ context + $[SEP] + $ target + $[SEP]$ is passed to a fully-connected layer.
    \item BERT-AEN \citep{Song_2019} that uses attentional encoder network with label smoothing regularization.
    \item The local context focus BERT (LCF-BERT, \citep{app9163389} where a local and global representation of the contexts are created through BERT that are further processed via multi-head self-attention.
\end{enumerate}
In addition, we also consider the R-GAT model that combines the power of BERT with Graph attention network and reported the best result so far for both the Laptop and Restaurants domain. 

\begin{table*}
\begin{center}
 \begin{tabular}{|c|c|c|c|c|c|c|} 
 \hline
 Model & \multicolumn{2}{c|}{Reported (no held out)} & \multicolumn{2}{c|}{Reproduced (no held out)} & \multicolumn{2}{c|}{Reproduced using 15\% held out} \\ [0.5ex] 
 \hline
  & Accuracy & F1 & Accuracy & F1 & Accuracy & F1 \\
 \hline
 ATAE-LSTM & 77.20 & - & 73.71 	& 55.87	& 73.29 (52.41)	& 54.59 (47.35) \\
 \hline
 RAM & 80.23 & 70.80 & 78.21 & 65.94 & 76.36 (59.29)	& 63.15 (56.36)
 \\
 \hline
 IAN & 78.60 & - & 76.80 & 64.24 & 76.52 (57.05) & 63.84 (55.11) \\ \hline
 BERT-SPC & 84.46 & 76.98 & 85.04* & 78.02* & $\mathbf{84.23}$ (68.84) & 76.28 (68.11) \\ \hline
 BERT-AEN & 83.12 & 73.76 & 81.73 & 71.24 & 80.07 (51.70) & 69.80 (48.97) \\ \hline
 LCF-BERT & 87.14 & 81.74 & $\mathbf{85.94}$ & $\mathbf{78.97}$ & 84.20* (69.38) & 76.28 (69.64) \\ \hline
 R-GAT+BERT	& 86.60	& 81.35 & 85.27 & 78.40 & 83.40 & 75.74 \\ \hline				
 STGNN-GRU & - & - & 84.93 & 77.65 & 83.66 (69.29) & 75.33 (68.45) \\ \hline
 STGNN-LSTM & - & - & - & - & 84.20* (\textbf{70.98}) & \textbf{76.55} (\textbf{70.44}) \\ \hline
 \end{tabular}
 \vspace{-2mm}
\caption{\label{table:restaurant}Comparison of predictions on the Restaurants dataset. The best results are highlighted in bold and * next to a number indicates the second best result.}
\end{center}
\end{table*}

\section{Results \& Analysis}
\vspace{-2mm}
For all the baseline models, it is difficult to know the exact hyperparameter settings in order to reproduce the results. Instead, we relied on the results that are obtained by \citep{10.1007/978-3-030-72240-1_7}. We have also included the originally reported results for the sake of completion and easy comparison. For all the datasets we have two sets of results, (a) the test set is used as a validation set and the model is decided based on the epoch with the best test set accuracy; and (b) 15\% of the training data is used as a validation set that decides the optimum number of epochs. Subsequently, the model is trained on the full train set till the optimum number of epochs and results are reported on the test set. For both the cases, average scores over 5 runs are reported for all the experiments. 

\subsection{Model Performance}
\vspace{-2mm}
Table~\ref{table:laptop} presents the results from the baseline models as well as our current model for the Laptop dataset. The first two columns show the originally reported test accuracy and F1-score without any held out validation data. The next two columns show the same metrics as obtained by \citep{10.1007/978-3-030-72240-1_7} again without any separate validation data. The last two columns show the same metrics with 15\% validation data (created from train set). As we can see the current method obtains the best result for both the accuracy and F1-score for this setup with a substantial improvement over the next best result from LCF-BERT. Our model also works well on the hard dataset with an improvement of 1.79 and 1.63 percent point, respectively, for the accuracy and F1-score. 

On the Restaurant dataset (Table~\ref{table:restaurant}) we show two different predictions from our model, one with GRU encoder and the second one with LSTM encoder. For GRU encoder, our model predictions are close to the best predictions of BERT-SPC and LCF-BERT while the gap in accuracy on the hard dataset is minimal. It is to be noted that the same set of hyperparameters is used in this case and not tuned specifically for the Restaurant dataset. Similar trend is also observed for the BERT-AEN and R-GAT models where the performance on the Laptop dataset is significantly better compared to the Restaurant dataset.
Using LSTM encoder, on the other hand, our model accuracy on the whole dataset is same as that of the best model whereas, on the hard dataset STGNN prediction outperforms the current best model. In case of F1 score, our model outperforms both on the overall and hard dataset. It is to be noted that on the Laptop dataset, the LSTM encoder based STGNN model does not perform better than the GRU based model. More on the choice of encoder is discussed later.

\begin{table*}
\begin{center}
 \begin{tabular}{|c|c|c|c|c|} 
 \hline
 Model & \multicolumn{2}{c|}{no held out} & \multicolumn{2}{c|}{using 15\% held out} \\ [0.5ex] 
 \hline
  & Accuracy & F1 & Accuracy & F1 \\
 \hline
 ATAE-LSTM & 83.13 & 55.98 & 81.65 (58.33) & 54.84 (39.25) \\ \hline
 RAM & 90.51 & 61.93 & 88.26 (83.33) & 59.67 (56.01) \\ \hline
 IAN & 87.58 & 59.16 & 87.41 (63.75) & 58.97 (42.85) \\ \hline
 BERT-SPC & 93.13 & 73.86 & 92.42 (89.58) & $\mathbf{73.83}$ (60.62) \\ \hline
 BERT-AEN & 88.69 & 72.25 & 87.54 (50.42) & 59.14 (32.96) \\ \hline
 LCF-BERT & 93.35 & 72.19 & 91.99 (91.67) & 72.13* (62.30) \\ \hline
 STGNN-GRU & $\mathbf{93.60}$ & $\mathbf{78.77}$ & 92.21* (90.0*) & 71.09 (60.90) \\ \hline
 \end{tabular}
\vspace{-2mm}
 \caption{\label{table:tshirt}Comparison of predictions on the Men's T-Shirt dataset. The best results are highlighted in bold and * next to a number indicates the second best result.}
\end{center}
\end{table*}

For the Men's T-Shirt and Television dataset all the previous results are reported by \citep{10.1007/978-3-030-72240-1_7}. Table~\ref{table:tshirt} shows the comparison for the Men's T-Shirt dataset where our model achieves the best results for the no held out scenario. For the 15\% validation data based case, the present model achieves competitive performance on the complete test data (a gap of only 0.2 percent point on accuracy). However, the gap increases to 1.67 percent point on the hard dataset. It is to be noted that there are only 48 examples in the hard test set. Similarly, on the Television dataset (shown in Table~\ref{table:tv}) our model achieves comparable results for both no held out and 15\% held out data. For the first case (no separate validation set) the gap in accuracy and F1-score with the best performing model (LCF-BERT) is 0.63 and 0.28 percent point, respectively. For the 15\% held out data, our model achieves the second best results with a gap of 0.21 and 0.56 percent points, respectively, on the accuracy and F1-score. Similarly, on the hard slice the gaps are also minimal at 0.4 percent point.   

\begin{table*}
\begin{center}
 \begin{tabular}{|c|c|c|c|c|} 
 \hline
 Model & \multicolumn{2}{c|}{no held out} & \multicolumn{2}{c|}{using 15\% held out} \\ [0.5ex] 
 \hline
  & Accuracy & F1 & Accuracy & F1 \\
 \hline
 ATAE-LSTM & 81.10 & 53.71 & 79.68 (53.92) & 52.78 (39.13) \\ \hline
 RAM & 84.29 & 58.68 & 83.02 (64.31) & 58.50 (50.07) \\ \hline
 IAN & 82.42 & 57.15 & 80.49 (54.31) & 56.78 (41.67) \\ \hline
 BERT-SPC & 89.96* & 74.68 & 88.56 (80.20) & 74.81 ($\mathbf{74.32}$) \\ \hline
 BERT-AEN & 87.09 & 67.92 & 85.94 (50.39) & 65.65 (38.08) \\ \hline
 LCF-BERT & $\mathbf{90.36}$ & $\mathbf{76.01}$ & $\mathbf{90.00 (80.98)}$ & $\mathbf{75.86}$ (73.72*) \\ \hline
 STGNN-GRU & 89.73 & 75.73* & 89.79* (80.59*) & 75.30* (73.32) \\ \hline
 \end{tabular}
 \vspace{-2mm}
\caption{\label{table:tv}Comparison of predictions on the Television dataset.}
\vspace{-2mm}
\end{center}
\end{table*}

\subsection{Error Analysis}
\vspace{-2mm}
We have also conducted a detailed analysis of the errors made by our model to understand if any discernible pattern exists. A summary of the distribution of the true class for different datasets are provided in Table~\ref{tab:error1}. It can be seen that most of the error is concentrated around the neutral class for the Laptop, Restaurant and Television dataset, whereas, for the Men's T-Shirt dataset the errors are uniform amongst the classes.

For the neutral classes the errors are broadly categorized into two classes: \vspace{-2mm}
\begin{itemize}
    \item \textbf{Presence of negation words}, examples: (a) "which it did not have , only 3 usb 2 ports .", (b) "no startup disk was not included but that may be my fault", (c) "there is no ""tools"" menu .", or (d) "the happy hour is so cheap , but that does not reflect the service or the atmosphere ."
    
    \item \textbf{Presence of negative/positive adjectives}, examples: (a) the only solution is to turn the brightness down, (b) "a lot of features and shortcuts on the mbp that i was never exposed to on a normal pc", (c) "premium price for the os more than anything else", or (d) "tiny restaurant with very fast service ."
\end{itemize}
while for the positive or negative true classes there are examples of general lack of understanding of the meaning due to their complexity or presence of double negation: 
\vspace{-2mm}
\begin{itemize}
    \item \textbf{Complicated}: (a) "if you ask me , for this \underline{price} it should be included", (b) "\underline{logic board} utterly fried , cried , and laid down and died", (c) "however , i can refute that \underline{osx} is `` fast '' .", or (d) "the sangria 's - watered down"
    \item \textbf{Double negation}: (a) screen - although some people might complain about low res which i think is ridiculous ., (b) i would have given it 5 starts was it not for the fact that it had windows 8 etc.
\vspace{-2mm}
\end{itemize}
In absence of enough training examples the onus of understanding the nuances of the language falls on the word/sentence representation, which also explains the relatively higher success rate of BERT.  

\begin{table}
\begin{center}
 \begin{tabular}{|c|c|c|c|c|}
 \hline
 Dataset & Positive & Negative & Neutral \\
 \hline
 Laptop & 32\% & 15\% & 53\% \\ \hline
 Restaurants & 25\% & 20\% & 55\% \\ \hline
 Men's Tshirt & 30\% & 36\% & 34\% \\ \hline
 Television & 33\% & 23\% & 43\% \\ \hline
 \end{tabular}
 \vspace{-2mm}
\caption{\label{tab:error1}Distribution of mis-prediction across different true classes}
\end{center}
\vspace{-2mm}
\end{table}

\section{Conclusion}
\vspace{-2mm}
We present a novel application of graph Fourier transform with spectral feature engineering hitherto limited to forecasting domain. The model learns an underlying graph structure from the raw data created by a BERT encoder. The advantage of this approach is that it does not require dependency parser based graph creation and thereby does not inherit any limitation of the parser. It is shown that the series of transformations involving GFT, DFT, convolution and GLU create powerful representations of the text resulting in the superior performance on SemEval-2014 datasets, namely "Laptop" and "Restaurants" domain. On the "Laptop" dataset we achieved the best results while on the "Restaurants" dataset our performance is at par with the current best prediction. On the recently released e-commerce datasets, our model performance is very competitive with a gap of 0.2-0.4 percent points. Although we have not done a full-scale hyper-parameter tuning, the effect of different components like the initial encoder and the final layer is studied. It is observed that the same set of hyper-parameters and architecture will not generate the best result across all the datasets. 

There are several possible future directions of work. If we view the current model as a spectral graph transformer that takes sequential input and generates sequential output there could be several other applications like, sequence tagging or natural language generation. Also, we have evaluated only BERT for sentence encoding and in future, other language models like RoBERTa and GPT can be explored.

\bibliography{anthology,acl2020}
\bibliographystyle{acl_natbib}

\appendix

\end{document}